\documentclass[12pt]{report}

\usepackage{graphicx}
\usepackage{subcaption}
\usepackage{float}
\usepackage{adjustbox}
\usepackage[utf8]{inputenc}
\usepackage{placeins}
\usepackage{amsmath}
\usepackage{booktabs}
\usepackage{siunitx}
\usepackage{caption}
\usepackage[strict]{changepage}
\usepackage[hidelinks]{hyperref}
\usepackage[protrusion=true,expansion=false]{microtype}
\usepackage{ragged2e}
\usepackage{amsmath}
\usepackage{amssymb}

\usepackage[numbers]{natbib}

\hypersetup{
  colorlinks=true,
  linkcolor=blue,
  citecolor=black,
  filecolor=magenta,
  urlcolor=blue,
}

\makeatletter
\def\@makechapterhead#1{%
  \vspace*{8\p@}
  {\parindent \z@ \raggedright \normalfont
    \ifnum \c@secnumdepth >\m@ne
        \huge\bfseries \@chapapp\space \thechapter
        \par\nobreak
        \vskip 12\p@
    \fi
    \interlinepenalty\@M
    \Huge \bfseries #1\par\nobreak
    \vskip 18\p@
  }}
\def\@makeschapterhead#1{%
  \vspace*{8\p@}
  {\parindent \z@ \raggedright
    \normalfont
    \interlinepenalty\@M
    \Huge \bfseries #1\par\nobreak
    \vskip 18\p@
  }}
\makeatother

\title{Automatic Brain Tumor Segmentation Using Deep Learning Methods}
\author{Mohsen Yaghoubi Suraki}
\date{2024}

\begin{document}

\maketitle

\thispagestyle{empty}
\vspace*{3cm}
\noindent\textbf{Note:} This is an unpublished research report prepared by the author in 2024. It is made publicly available here as a record of the author's independent research work.\\[0.5em]
\noindent \copyright\ 2024 Mohsen Yaghoubi Suraki
\newpage

\tableofcontents
\newpage

\chapter*{Abstract}
\addcontentsline{toc}{chapter}{Abstract}

Glioma is a harmful brain tumor that requires early detection to ensure better health results. Early detection of this tumor is key for effective treatment and requires an automated segmentation process. However, it is a challenging task to find tumors due to tumor characteristics like location and size. A reliable method to accurately separate tumor zones from healthy tissues is deep learning models, which have shown promising results over the last few years. In this research, an Adaptive Dual Residual U-Net with Attention Gate and Multiscale Spatial Attention Mechanisms (ADRUwAMS) is introduced. This model is an innovative combination of adaptive dual residual networks, attention mechanisms, and multiscale spatial attention. The dual adaptive residual network architecture captures high-level semantic and intricate low-level details from brain images, ensuring precise segmentation of different tumor parts, types, and hard regions. The attention gates use gating and input signals to compute attention coefficients for the input features, and multiscale spatial attention generates scaled attention maps and combines these features to hold the most significant information about the brain tumor. We trained the model for 200 epochs using the ReLU activation function on BraTS 2020 and BraTS 2019 datasets. These improvements resulted in high accuracy for tumor detection and segmentation on BraTS 2020, achieving dice scores of 0.9229 for the whole tumor, 0.8432 for the tumor core, and 0.8004 for the enhancing tumor.

\chapter*{Acknowledgements}
\addcontentsline{toc}{chapter}{Acknowledgements}

Thanks to everyone who supported me along the way.


\chapter{Introduction}

In the interdisciplinary fields of computer science and medical imaging, the application of artificial intelligence has become a pivotal element for breakthroughs in the diagnosis of brain tumors. This tumor, which is an abnormal growth of cells, poses a significant threat to the patient's health and survival. Given the complexity and variability of brain tumors like size, location, and malignancy, precise and early detection is not just a medical necessity but a technological challenge.
Brain tumors can be detected through medical imaging, and among the medical imaging technologies, magnetic resonance imaging (MRI) has created new opportunities in healthcare diagnostics ~\cite{weller2017european}. There are four modalities in MRI images for tumor: fluid-attenuated inversion recovery (FLAIR), T1 weighted (T1), T2 weighted (T2), and contrast-enhanced T1 weighted (T1ce). Fig 1.1 shows different modalities along with their ground truth.

    \begin{figure}[H]
    \centering
     \includegraphics[width=\textwidth, keepaspectratio]{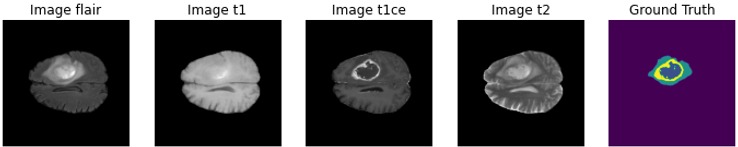}
    \caption[Different Modalities of Brain MRI with their Ground Truth]{Display of four brain MRI modalities (FLAIR, T1, T1ce, T2) on the left alongside the corresponding Ground Truth segmentation on the right}
    \label{fig:modalities}
    \end{figure}

The landscape of tumor detection and segmentation has traditionally depended on manual interpretation of medical images. However, these methods are often time-consuming and possibly subject to human error. This highlights the need for automated, accurate, and efficient techniques. Alongside this, the field of computer science has witnessed a significant increase in the capabilities of machine learning, especially with the advent of deep learning techniques. Among various deep learning architectures, convolutional neural networks (CNNs) have emerged as particularly effective in image recognition and segmentation tasks ~\cite{havaei2017brain}. However, these methods typically employ additional convolutional layers and pooling layers, which can lead to a network degradation problem. This often results in reducing segmentation accuracy and the overall performance of the model.
To address this challenge, various techniques have been explored, such as skip connections ~\cite{ronneberger2015u}, residual connections ~\cite{he2016deep}, and attention mechanisms ~\cite{vaswani2017attention}. These methods aim to alleviate the negative effects of network degradation and maintain or even enhance segmentation precision. Residual networks, for instance, were introduced as a solution to increase the performance of deep neural networks ~\cite{he2016deep}. Alongside these developments, the U-Net Architecture has become a widely adopted framework in this field ~\cite{ronneberger2015u}. This research proposes a novel 3D-Net model to segment a brain tumor. This enhanced model integrates a new architecture of advanced attention gates, multiscale spatial attention, and two ResNet blocks with adaptive skip connection into the traditional U-Net framework. Our experiments aim to demonstrate that this model offers improved performance over existing architectures in brain tumor segmentation. The contributions of this study are summarized as follows:
\begin{itemize}
\item In this research, an adaptive dual residual block is introduced for nuanced feature extraction from brain MRI scans. This sequential residual block helps to distinguish high-level semantics and intricate details crucial for precise segmentation.
\item Our model employs a multi-scale spatial attention mechanism with varied convolutional kernel sizes (3x3, 5x5, 7x7) to create scaled attention maps. This method enhances tumor segmentation accuracy by focusing on critical spatial features across scales.
\item The refined attention gate mechanism also utilizes sequential processing and group normalization, offering sophisticated feature modulation.
\item Evaluated on 2019, and 2020 Brain Tumor Segmentation (BraTS) datasets ~\cite{menze2014multimodal}, our model showcases remarkable segmentation improvements, validated by enhanced Dice scores.
\end{itemize}

\chapter{Literature Review}

The field of brain tumor segmentation has experienced substantial advancements and deep learning methods have revolutionized the field of medical imaging, particularly CNNs. These algorithms excel at automatically learning hierarchical feature representations from medical images for tasks such as image classification, and image segmentation. In this section, we delve into the literature reviews of Semantic Brain Tumor Segmentation, exploring the various aspects of deep learning modules employed in our study.

\vspace{-1.5mm}
\section{Semantic Brain Tumor Segmentation}
Semantic segmentation plays a pivotal role in medical imaging by delineating specific structures and regions within medical scans. It refers to the process of partitioning an image into several segments with the aim of simplifying its representation into something more meaningful and easier to analyze. In the context of medical imaging, semantic segmentation helps in identifying and classifying pathology, anatomical structures, or regions of interest ~\cite{havaei2017brain}.
Glioma is a brain tumor that originates in glial cells, and it represents the majority of primary brain tumors ~\cite{goodenberger2012genetics}. Tumors are classified on a scale from I to IV, with grade I tumors growing slowly and grade IV tumors growing quickly. A grade I tumor is deemed benign, while a grade IV tumor is considered malignant. This classification system for tumors was created by the World Health Organization in 2007 and updated in 2016 ~\cite{louis20162016}. In a medical context, a simpler grading system is often applied to gliomas, where grades I to II are categorized as Low-Grade Glioma (LGG) and grades III to IV as High-Grade Glioma (HGG). However, this tumor is divided into three distinct groups in the BraTS dataset ~\cite{menze2014multimodal}, which is used in this research. Key types of tumor tissues that are commonly segmented include necrosis, edema, and enhancing tumors. Fig 1.2 shows these annotations.

    \begin{figure}[H]
    \centering
     \includegraphics[width=0.8\textwidth, keepaspectratio]{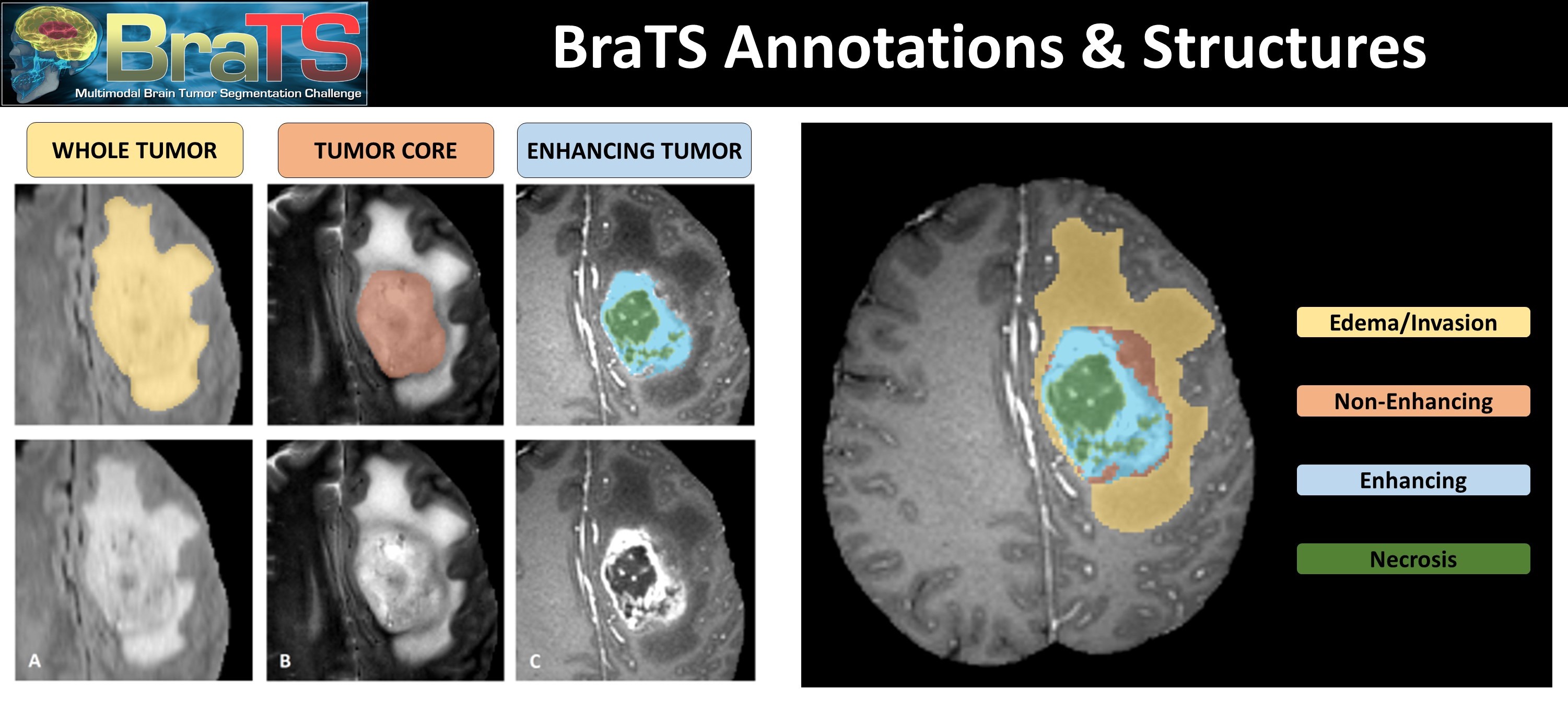}
    \caption[Multimodal Brain Tumor Segmentation Annotations]{Illustration of Brain Tumor Segmentation and Annotations. The left side displays MRI scans highlighting the whole tumor, tumor core, and enhancing tumor regions. The rightmost image presents a detailed annotation of tumor components, including edema/invasion, non-enhancing, enhancing, and necrosis regions as per the BraTS standards. ~\cite{menze2014multimodal}}
    \label{fig:brainsubregions}
    \end{figure}

On the right side, the different glioma sub-regions as identified on MRI scans; Edema denotes areas experiencing fluid accumulation, generally surrounding necrotic or rapidly growing tumor zones, manifesting as swelling or fluid retention, Non-enhancing refers to a portion of the tumor that does not show an increased signal intensity after the administration of a contrast agent (gadolinium), Enhancing Tumor shows active cellular proliferation and it factors critical for malignancy determination, Necrosis: it is identified in tumor segments where cell death has occurred, typically central within the tumor, reflecting areas of uncontrolled cellular demise. On the left side, there are three specific sub-regions for evaluation: the enhancing tumor (ET), the tumor core (TC), and the whole tumor (WT). Enhancing Tumor (ET) is the region that shows hyper-intensity on T1-weighted post-contrast images (T1Gd) compared to both the T1 images and normal white matter on T1Gd images. It is the active tumor area that takes up contrast agent.
Tumor Core (TC) is the central area of the tumor, typically targeted for surgical resection. The TC includes the ET, as well as the necrotic (NCR, fluid-filled) and non-enhancing (NET, solid) parts of the tumor. Whole Tumor (WT) is the region that encompasses the full extent of the disease, combining the TC and the peritumoral edema (ED). The ED is usually characterized by a hyper-intense signal on Fluid-attenuated inversion recovery (FLAIR) MRI scans.
Traditional methods, such as registration-based approaches ~\cite{campadelli2009liver} and atlas-based techniques ~\cite{isgum2009multi}, have laid the groundwork for medical image segmentation by utilizing statistical shape models and multi-atlas label fusion. In ~\cite{garcia2020review}, a comparative study presents an overview of deep learning approaches against traditional methods. The development and application of U-Net for medical image segmentation is an example of how deep learning methodologies have transformed the field, especially in terms of efficiency and accuracy ~\cite{ronneberger2015u}.

\vspace{-1.5mm}
\section{Convolutional Neural Networks (CNNs)}
Convolutional Neural Networks (CNNs) form the backbone of most deep learning approaches to medical image analysis. Introduced in the 1980s by Fukushima ~\cite{fukushima1980neocognitron} and popularized by LeCun et al ~\cite{lecun1998gradient}. CNNs could extract intricate patterns from medical images, such as MRIs or CT scans, learning features and patterns relevant to diseases, anomalies, and physiological structures ~\cite{litjens2017survey}. Research by Havaei et al. ~\cite{havaei2017brain} highlighted the adaptability of CNNs in accurately segmenting brain tumors and Akkus et al. further highlighted the pivotal role of CNNs in advancing brain MRI segmentation ~\cite{akkus2017deep}. The brain tumor segmentation (BraTS) benchmark, discussed by Menze et al. ~\cite{menze2014multimodal}, serves as a vital tool for evaluating CNN segmentation models. Additionally, Kamnitsas et al presented an advanced multi-scale 3D CNN that significantly improves brain lesion segmentation ~\cite{kamnitsas2017efficient}. The application of CNNs in brain tumor segmentation has been marked by continuous innovation, with architectures like U-Net ~\cite{ronneberger2015u} and its variants being specifically tailored for medical imaging challenges. These models adapt the foundational principles of CNNs to address the unique complexities of tumor segmentation.

\section{U-Net Architecture}

A significant leap in medical image segmentation was the introduction of the U-Net architecture. U-Net is a type of CNNs that has been optimized for biomedical image segmentation tasks and provide fast and precise image segmentation ~\cite{ronneberger2015u}. The novelty of U-Net lies in its symmetrical expanding path, which complements the traditional contracting path of its CNN layers. The contracting path, akin to a standard CNN, decreases the spatial size of the input while increasing the depth. This characteristic helps in capturing the context from the input images. Conversely, the expanding path allows for upsampling of feature maps and the recovery of spatial information, facilitating precise pixel-level predictions.
Another unique feature of the U-Net architecture is its skip connections. These connections enable information transfer between layers of the same size in the reducing path to their corresponding layers in the enlarging path. This element facilitates the network in leveraging both local and global contexts to achieve improved segmentation results.

\section{Residual Networks}

Residual Networks, or ResNets ~\cite{he2016deep}, address the problem of training deep neural networks by using skip connections or shortcuts, known as "residual blocks". These links enable the model to establish a function that guarantees each subsequent layer will operate at least as effective as the preceding one, helping to reduce the issue of diminishing gradients. The main idea is to redefine the underlying mapping to be learned by the network, making optimizing and enabling the training of deeper networks easier. Adding these residual connections to U-Net can help learn more complex mappings and make optimization easier, thus potentially improving the model's performance on complex segmentation tasks.
Despite the significant achievements of ResNets, they are not flawless. In particular, their performance on segmentation tasks, including brain tumor segmentation, can be improved further. In ~\cite{de2021deep}, a system utilizing an 18-layer ResNet architecture is proposed for identifying brain tumor types. Another study ~\cite{saha2021brain}, introduces a multipathway architecture built upon the U-Net with residual connections, which efficiently predicts multiple tumor types in a single pass. Furthermore, a paper ~\cite{ramasamy2023multi} presents a model that combines a modified LinkNet structure with the ResNet152 for advanced tumor segmentation. While ResNets provides a powerful tool for deep learning tasks, their potential can be further unleashed through the right combination of its methods and architectures. Fig 2.2 describes two distinct neural network block diagrams ~\cite{zhang2023dive}. The left one is the Regular Block. It starts with an input labeled "x," which is passed through a "Weight layer." After this layer, the result is processed by an "Activation function." The final outcome from this sequence is denoted as "f(x)." The second diagram represents the Residual Block. Like the regular block, it initiates with an input labeled "x" which traverses through a "Weight layer" and then an "Activation function." A distinctive feature of the residual block is the addition of the original input "x" directly to the output of the second activation function. This addition of the original input to the final output differentiates the residual block from the regular one, enabling the network to learn residual or difference functions more efficiently.
    \begin{figure}[H]
    \centering
    \includegraphics[width=0.6\textwidth, keepaspectratio]{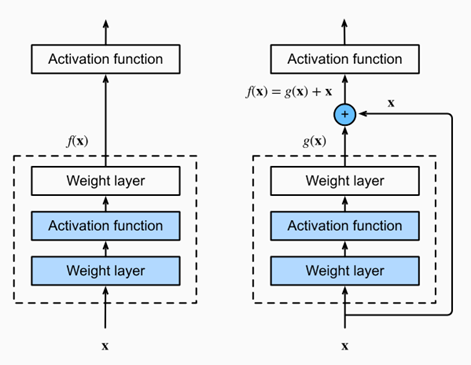}
    \caption[Neural network diagrams: a regular block (left) and Residual block (right)]{Comparison of neural network blocks: Standard vs. Residual. The standard block directly learns the desired mapping $f(x)$, while the residual block learns the residual $g(x) = f(x) - x$, facilitating easier learning of the identity function $f(x) = x$. ~\cite{zhang2023dive}}
    \label{fig:residual}
    \end{figure}

\section{Attention Mechanisms}
Attention mechanisms initially presented in Natural Language Processing (NLP) ~\cite{bahdanau2014neural} and have gained popularity in many other areas, including computer vision ~\cite{vaswani2017attention}. In the landscape of attention mechanisms, several innovative approaches such as Transformer models ~\cite{vaswani2017attention}, Convolutional Block Attention Module (CBAM) ~\cite{woo2018cbam}, and Squeeze-and-Excitation (SE) networks ~\cite{hu2018squeeze} have emerged, and among these attention mechanisms, we select Attention Gate (AG) ~\cite{oktay1804attention} and Spatial Attention ~\cite{woo2018cbam} for their unique attributes suited to the challenges of brain tumor segmentation. These methods demonstrated effectiveness in medical imaging tasks and their computational efficiency, which is crucial for processing the large volumes of data typical in this domain ~\cite{xie2023attention}.

\subsection{Attention Gates}
\justify
Attention Gate (AG) is a type of attention mechanism that allows the network to concentrate on distinct areas of the input image ~\cite{schlemper2019attention}. Fig 2.3 depicts an attention gate (AG) module designed to enhance feature representation selectively by focusing on relevant information within an input. This process involves element-wise multiplication of the input features with attention coefficients, followed by a resampler to align the attended features back onto the original shape ~\cite{oktay2018learning}. The gating path can improve the model's performance by emphasizing relevant features and diminishing the irrelevant ones.
\FloatBarrier
    \begin{figure}[htbp]
    \centering
     \includegraphics[width=\textwidth, keepaspectratio]{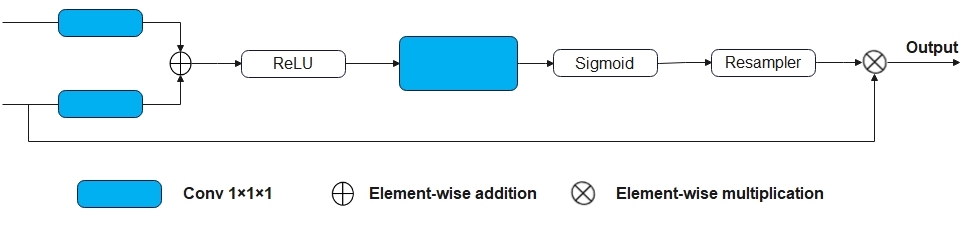}
    \caption[The Attention Gate Schematic Module]{Diagram of the Attention Gate Module, showing the selective focus mechanism on important input features through attention coefficients and subsequent feature refinement for enhanced model performance ~\cite{oktay2018learning}}
    \label{fig:attentiongate}
    \end{figure}
\FloatBarrier
In recent years, the integration of the Attention Gate mechanism into the U-Net architecture has become a prevalent approach for brain tumor segmentation. One study ~\cite{huang2021gcaunet} introduced the GCAUNet model, which accentuates tumor details through a detail recovering path, enriches key features using a group cross-channel attention module, and captures a multiscale context with a multiscale input path. Another study ~\cite{maji2022attention} presented the Attention Res-UNet with Guided Decoder (ARU-GD). This innovative deep learning architecture combines a guided decoder with attention gates, generating advanced feature maps that selectively activate relevant information. A separate study ~\cite{chinnam2022multimodal} unveiled the Multimodal Attention-gated Cascaded U-Net (MAC U-Net). Specifically tailored for the detection and segmentation of early-stage low-grade brain tumors, this model integrates group normalization, attention gates, and skip connections to enhance precision. Further exploring the potential of attention mechanisms, another study ~\cite{zhang2020attention} introduced the AGResU-Net. This model seamlessly merges residual modules and attention gates within the foundational U-Net framework, making it especially effective for segmenting smaller tumors.

\subsection{Spatial Attention}
Spatial attention allows the network to focus on pertinent sections of the image. However, this attention mechanism works by assigning varying importance levels to different spatial locations ~\cite{woo2018cbam}. This capability can assist the model in concentrating on the image regions that hold the most details about the target. Figure 2.4 illustrates the spatial attention module, which processes a feature map through pooling operations before passing it through a convolutional layer ~\cite{woo2018cbam}. The resulting spatial attention map emphasizes specific areas of the input feature map, thereby enhancing feature extraction for subsequent layers of the neural network. Using multiscale spatial attention offers a nuanced approach to analyzing images across various scales. This mechanism enables models to dynamically adjust their focus, emphasizing different spatial regions and scales within an image to capture both granular details and broader contextual information. For instance, the MDA-Unet introduces a sophisticated multi-scale spatial attention module that leverages spatial attention maps from a hybrid hierarchical structure ~\cite{amer2022mda}. Similarly, MSCSA-Net employs a blend of local channel spatial attention and multi-scale strategies to improve semantic segmentation of complex remote sensing images~\cite{liu2023mscsa}. These advancements highlight the growing role of attention mechanisms in refining image segmentation models.

\begin{figure}[H]
\centering
\includegraphics[width=0.58\textwidth, keepaspectratio]{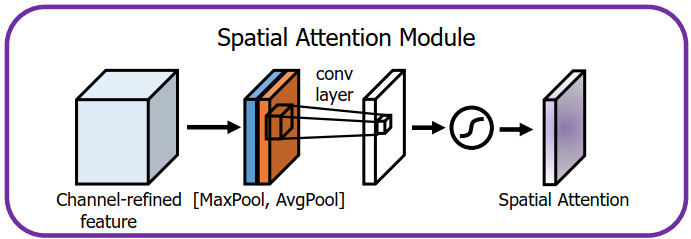}
\caption[The Spatial Attention Module]{The Spatial Attention Module diagram, illustrating the process where the network assigns variable importance to different areas of the input, enhancing the focus on regions with critical information by refining features through pooling and convolutional layers. ~\cite{woo2018cbam}}
\label{fig:spatialattention}
\end{figure}

A variety of novel architectures have been proposed with the integration of Spatial attention mechanism. One such development is presented in a study~\cite{mazumdar2022fully} where the authors introduced the Efficient Spatial Attention Network (ESA-Net). This model, an optimized variant of the renowned U-Net, incorporates the innovative Efficient Spatial Attention (ESA) blocks. These blocks, made up of depthwise separable convolution layers coupled with a spatial attention module, are designed to bolster both efficiency and accuracy. Furthering the exploration of attention mechanisms, another study ~\cite{chi2022scar} showcased the SCAR U-Net, a 3D residual U-Net architecture. Distinguished by its integration of channel and spatial attention processes, this model also employs residual blocks and leverages a complex loss function. With attention gates strategically positioned between its sampling blocks and a range of data preprocessing techniques, the SCAR U-Net is tailored for effective brain tumor segmentation. Another contribution to this field is outlined in a study ~\cite{nie2020spatial}, where the SA2FNet was unveiled. This 3D spatial attention-centric network targets the challenge of inadequate feature fusion. By harnessing spatial attention mechanisms, the SA2FNet meticulously refines and amalgamates low-level and high-level features, resulting in enriched segmentation details. Additionally, the inclusion of deep supervision ensures accelerated training convergence, thereby enhancing the model's ability to discern features in its early stages.

\section{Dataset Description}

In this research, we utilize the BraTS 2020 dataset. This data was compiled for the Multimodal Brain Tumor Segmentation Challenge (BraTS) ~\cite{jadon2020survey}. This dataset comprises four 3D MRI modalities of brain scans for 369 patients with Glioma. Each modality responds differently to the water content of the brain, and each 3D scan features a dimension of 240 x 240 x 155 slices, offering a comprehensive view of the brain's structure.
The BraTS 2020 dataset includes four labels: Label 0 (Background), Label 1 (Necrosis and non-enhancing tumor), Label 2 (Edema), and Label 4 (Enhancing tumor). Each of these labels corresponds to unique characteristics of the brain scans, which are individually marked by one to four neuroradiologists. For the purpose of this study, labels 1, 2, and 4 were transformed into separate binary segmentation masks. Building on our primary training, we further experimented with BraTS 2019 datasets ~\cite{menze2014multimodal}~\cite{bakas2017advancing}~\cite{bakas2018identifying}. The BraTS 2019 dataset introduced a training set of 335 glial tumor cases. Its validation data, comprising 125 cases.

\chapter{Problem Statement}

Brain tumor segmentation in medical imaging is a critical yet challenging task in the realm of computer-aided diagnosis and treatment planning. One of the primary challenges in brain tumor segmentation is the inherent variability of tumors. Tumors vary greatly in size, shape, location, and even within the same type of tumor, like gliomas ~\cite{louis20162016}. This variability makes it difficult to design a one-size-fits-all algorithm for tumor segmentation. As a result, segmentation models often suffer from a lack of generalizability when applied to diverse datasets.
While deep learning models, especially convolutional neural networks (CNNs), have shown great promise in this domain, they are not without their limitations. A common issue is the requirement of large amounts of labeled data. Annotating medical images for training these models is time-consuming and requires expert knowledge, making it a costly and labor-intensive process ~\cite{litjens2017survey}.
The quality and availability of medical imaging data pose another significant challenge. High-quality, annotated datasets like those from the BraTS challenge are limited, and there's a lack of standardized protocols across different institutions and regions, leading to variations in imaging data quality. This inconsistency can severely affect the performance of deep learning models trained on specific datasets ~\cite{bakas2017advancing}.
\section{Research Objectives}
Given these challenges, the primary objective of this research is to develop an advanced deep learning-based framework for brain tumor segmentation that addresses the issues of variability in tumor presentation and limitations in current models. To satisfy this objective, it is important to answer this question "To what extent can a deep learning-based framework, enhanced with attention mechanisms and ResNet improve the generalizability, efficiency with limited annotations, accuracy, and adaptability in brain tumor segmentation?" The research aims to:
\begin{itemize}
\item Develop a 3D U-Net architecture enhanced with multiscale spatial attention and attention gates. This architecture aims to focus the model's learning on relevant features across different scales, improving segmentation accuracy. The multiscale spatial attention mechanism allows the model to prioritize areas of interest within the images, addressing the challenge of tumor variability. Attention gates are integrated to refine the model's focus further, improving the precision of segmentation across varying tumor presentations.
\item Apply a stratified division of the dataset to ensure a fair and representative distribution of tumor types (NET, ED, ET) and sizes across training, validation, and testing sets. This strategy is intended to mitigate sampling bias and improve the model's generalization capability across varied tumor presentations. The methodology involves Computing tumor size, and voxel counts for each subtype within patient data and categorizing tumor sizes and subtypes to facilitate stratified sampling. We used 5 Fold for cross-validation while preserving the proportion of each category across the splits.
\item Utilize segmentation performance metrics such as Dice coefficient, Hausdorff distance, Sensitivity, and Specificity to evaluate the accuracy and reliability of the segmentation. The statistical significance of improvements observed in these metrics will be evaluated using paired t-tests, verifying if the proposed model's improvements are statistically significant and not due to chance. Additionally, calculating Cohen's d will quantify these improvements' effect sizes, providing context on their real-world impact.
\item Benchmark the developed model against existing state-of-the-art models on the same datasets to highlight advancements in segmentation accuracy, generalizability, and efficiency with limited annotations.
\end{itemize}

The successful development of a robust and reliable brain tumor segmentation tool has the potential to significantly impact clinical practices. It can aid in accurate diagnosis, facilitate personalized treatment planning, and potentially improve patient outcomes in brain tumor cases. Moreover, this research could pave the way for further advancements in applying deep learning in medical imaging, beyond brain tumors, to other areas of diagnostic imaging.

\chapter{Proposed Method}
Despite of many advantages of using U-Net in segmentation, it exhibits certain limitations for more complex tasks such as brain tumor segmentation, particularly when the regions of interest in an image are small or subtle ~\cite{menze2014multimodal}. To overcome these challenges, several enhancements have been proposed to address U-Net's limitations. One of the significant adaptations is the development of a 3D U-Net, which extends U-Net's 2D operations into 3D, thereby enabling more effective handling of volumetric images ~\cite{cciccek20163d}. This variation has proved particularly useful in tasks such as brain tumor segmentation, which utilizes three-dimensional MRI scans. Furthermore, the integration of residual connections and attention mechanisms has shown promising results. Residual U-Net, for instance, merges the benefits of U-Net and Residual Network (ResNet), enabling deeper networks without the issue of vanishing gradients ~\cite{he2016deep}.
On the other hand, Attention U-Net uses attention gates to selectively focus on relevant features and suppress irrelevant ones, potentially improving segmentation performance in complex tasks ~\cite{schlemper2019attention}. While U-Net has established a significant foundation for biomedical image segmentation, there still exists potential for further improvements. By embracing approaches such as residual learning, attention mechanisms, and complex connection structures, the proficiency and versatility of biomedical image segmentation can steadily progress.
The first significant innovation in the Adaptive Dual Residual U-Net with Attention Gate and Multiscale Spatial Attention Mechanisms (ADRUwAMS) is the integration of adaptive dual residual connections in the form of Residual Blocks (ResBlocks). These ResBlocks form the backbone of the encoder part of the model, replacing the traditional convolutional layers used in U-Net models. Each ResBlock incorporates two 3D convolutional layers, utilizing ReLU activations and Group Normalization for effective feature extraction. The residual connection within each block allows the model to bypass the main layers if necessary, alleviating potential issues of vanishing or exploding gradients and enabling the model to learn identity functions. Consequently, the ResBlocks helps capture both simple and complex patterns in the data across different levels of abstraction.
The ADRUwAMS's main strength comes from its advanced attention mechanisms, which include both spatial attention and attention gates. The spatial attention blocks apply a multiscale 3D convolution to the input and use a sigmoid activation function to generate a spatial attention map. These different scaled attention maps are added together and then multiplied with the original input to direct the model's attention toward the regions of extracted features that hold the most significant information. Meanwhile, the attention gates use gating and input signals to compute attention coefficients for the input features. This gate guides the network to give more emphasis to certain channels and suppress others, facilitating the model's ability to identify intricate and varied dependencies in the 3D input data. The proposed architecture is illustrated in Figure 4.1.
\FloatBarrier
    \begin{figure}[htbp]
    \centering
     \includegraphics[width=\textwidth, keepaspectratio]{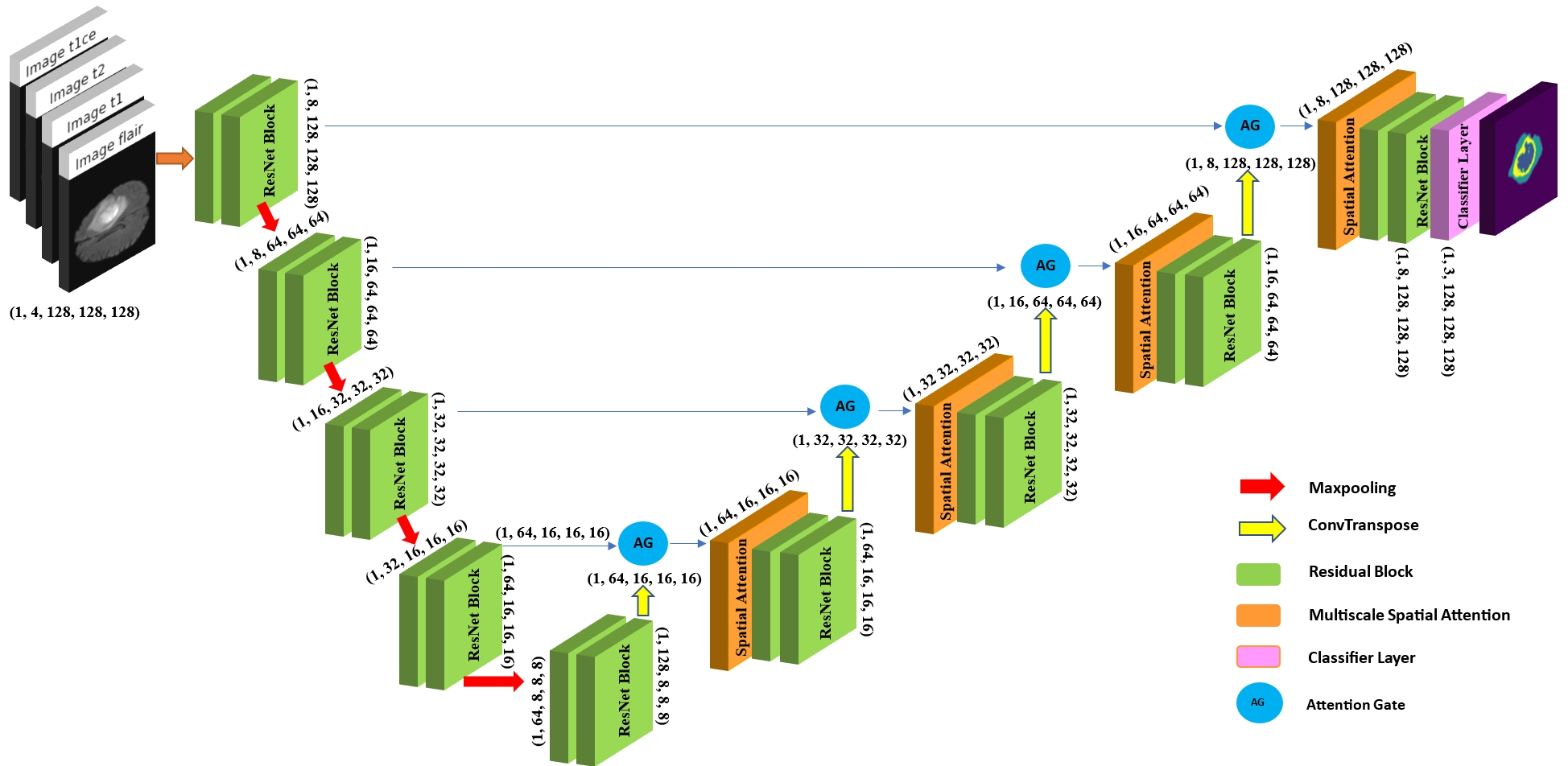}
    \caption[The architecture of ADRUwAMS]{Depiction of the Adaptive Dual Residual U-Net with Attention Gate and Multiscale Spatial Attention Mechanisms (ADRUwAMS), showcasing the integration of advanced attention features for enhancing salient features across multiple scales in 3D input data.}
    \label{fig:proposedmethod}
    \end{figure}
\FloatBarrier

The model's structure is designed to extract and enhance relevant features from images through a sophisticated downsampling and upsampling process. Initially, the downsampling path employs encoder blocks that sequentially double the feature maps, beginning with eight. Each encoder block comprises two residual blocks and a max-pooling layer, crucial for capturing the contextual details of the input image. This is followed by a bottleneck stage that processes the features without further downsampling. Then, in the decoding phase, a transposed 3D convolution expands the feature maps, which are then filtered through an attention gate to highlight salient features and suppress the less relevant ones. Additionally, a spatial attention mechanism applies varying weights across the feature map to spotlight areas of interest. This allows the model to focus on the most informative regions. After this attention-focused processing, two residual blocks refine the features at each decoding stage, which restores the spatial resolution step by step until the output is fully reconstructed. In the final step, a convolution layer of 1×1×1 dimensions is utilized to transform the multi-channel feature maps to the requisite number of classes, producing the ultimate segmentation map. In this case, the output size is (1, 3, 128, 128, 128), where 3 represents the number of segmentation classes. The proposed model employs an encoder-decoder architecture with residual connections, attention gates, and spatial attention mechanisms. It shows the potential for accurate volumetric segmentation by combining the context information from the encoder pathway and the localization cues from the decoder pathway, which are further enhanced by the attention mechanisms.

In the dual ResNet block, the output of one block serves as the input to the subsequent block and the shortcut connection in a ResNet block is used to add the input X to the output of the ResNet block's layers, which can either be an identity mapping or a convolutional transformation to match the dimensions. For a given input tensor X to the ResNet block, the shortcut connection Y is defined as follows:

\begin{equation}
Y =
\begin{cases}
X, & \parbox{10cm}{if the number of input channels \(C_{in}\) equals the number of output channels \(C_{out}\)}\\
GN(Conv_{1 \times 1 \times 1}(X)), & \text{otherwise}
\end{cases}
\end{equation}

Where GN stands for Group Normalization and \( Conv_{3 \times 3 \times 3} \) is a convolution operation with a filter size of \( 3 \times 3 \times 3 \) which is used to match the number of channels of the input \( X \) to the dimensions required for element-wise addition with the output of the ResNet block's layers. The condition checks if the number of input channels \( C_{in} \) is the same as the number of output channels \( C_{out} \). If the condition is met, the shortcut is an identity mapping; if not, a convolutional layer followed by normalization is applied to adjust the dimensions accordingly. This shortcut connection \( Y \) is then element-wise added to the output of the sequential convolutional layers within the ResNet block to form the final output of the ResNet block, \(F_{\text{ResNet}}\). The ResNet Convolutional Layers and the ResNet Block Output are as follows:

\begin{equation}
F_1 = GN\left(Conv_{3\times3\times3}\left( ReLU\left( GN\left(Conv_{3\times3\times3}(X)\right)\right)\right)\right)
\end{equation}

\begin{equation}
F_{\text{ResNet}} = ReLU(F_1 + Y)
\end{equation}

\(F_{\text{ResNet}}\) will be the input tensor to the Attention Gate, which can be \(G\) as gating signal if comes from encoder and \(X_{dec}\) if it comes from decoder layer. The transformation operations for both two inputs are as follows:

\begin{equation}
G_{\text{trans}} = ReLU\left( GN\left(Conv_{1\times1\times1}(G)\right)\right)
\end{equation}

\begin{equation}
X_{\text{trans}} = ReLU\left( GN\left(Conv_{1\times1\times1}(X_{dec})\right)\right)
\end{equation}

Then the transformed gating signal and features are then combined, and Attention coefficients \(\psi\) are computed by applying a sigmoid activation to a convolved output.

\begin{equation}
\psi = \sigma(\text{Conv}_{1\times1\times1}(G_{\text{trans}} + X_{\text{trans}}))
\end{equation}

Then the attention-modulated feature map \( F_{\text{Attn}} \) is then obtained by:

\begin{equation}
F_{\text{Attn}} = X_{\text{dec}} \otimes \psi
\end{equation}

The attention-modulated feature map then passes through a multiscale spatial attention mechanism that aggregates information across various scales to capture both fine and coarse details:

\begin{equation}
A_k = \sigma(\text{Conv}_{k\times k\times k}(F_{\text{Attn}})) \text{ for } k \in \{3,5,7\}
\end{equation}

For each kernel size \( k \) in the set \( \{3, 5, 7\} \), the feature map \( F_{\text{Attn}} \) is convolved to produce an attention map, \( A_k \), followed by a sigmoid activation. These attention maps from different scales are combined to form a single multiscale attention map \( S \), which is the sum of all \( A_k \).

\begin{equation}
S = \sum_{k \in \{3,5,7\}} A_k
\end{equation}

The final output, \( F_{\text{MSA}} \), is the element-wise product of \( F_{\text{Attn}} \) and the combined attention map \( S \), thus refining the feature map with contextually relevant information across multiple spatial hierarchies.

\begin{equation}
F_{\text{MSA}} = F_{\text{Attn}} \odot S
\end{equation}

A schematic diagram of the proposed architecture is shown in figure 4.2. The proposed architecture employs ResNet blocks as foundational components for feature extraction. Each block processes the input through a series of convolutions, where the kernel size is adjusted to the specific layer's requirements. To maintain the integrity of the input's spatial dimensions, a skip connection is employed. To enhance the skip connection, an advanced attention gate is introduced, which uses sequential processing of performing convolution operation for both gating and input signals, with intermediate group normalization and following activation function. Group normalization offers benefits in situations where batch size is small or inconsistent. In addition to the attention mechanism, multiscale spatial attention is integrated into the model to increase the model's performance. It involves processing 3D Convolution block outputs into attention maps using a Sigmoid Activation block. These maps are then element-wise multiplied with the input tensor, resulting in a weighted tensor with spatially focused features. Finally, the weighted tensor as the output contains the spatially attended features for the next decoder layer.
\FloatBarrier
    \begin{figure}[htbp]
    \centering
     \includegraphics[width=\textwidth, keepaspectratio]{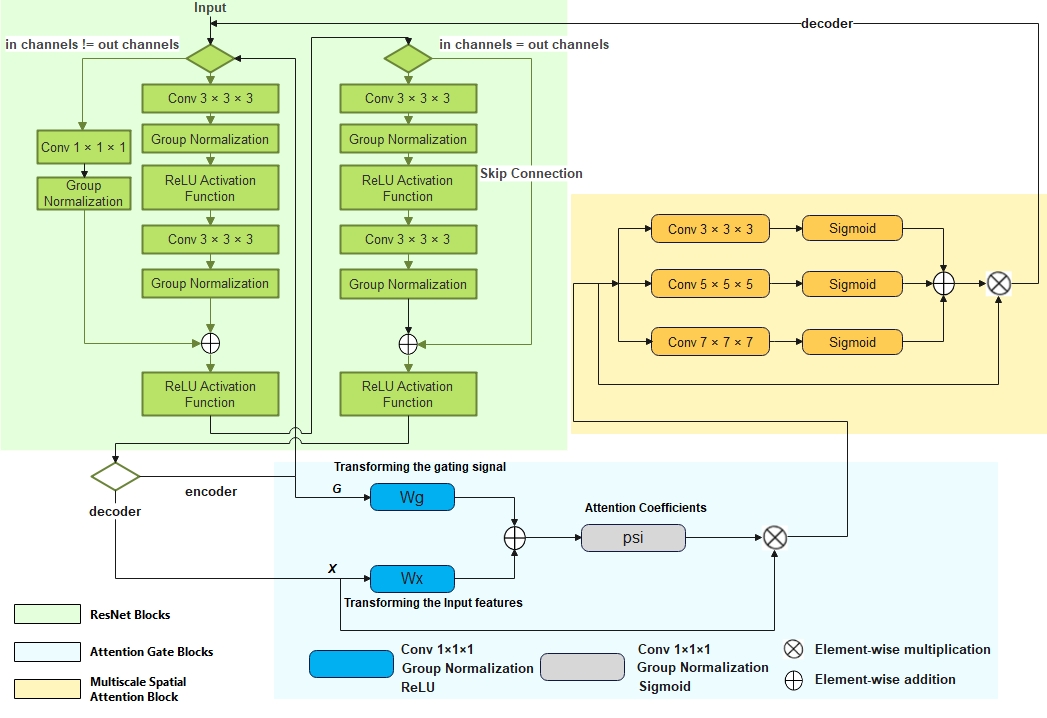}
    \caption[ResNet with Integrated Attention Mechanisms]{Depiction of a Residual Neural Network (ResNet) with Integrated Attention Mechanisms. The schematic outlines ResNet blocks as the foundation for feature extraction, supplemented by skip connections for spatial integrity, and group normalization for consistent learning. Attention modules further refine the output by transforming the input with 3D convolutional maps and a sigmoid function to direct the model's learning towards significant areas, resulting in a carefully weighted output prepared for the subsequent decoding phase}
    \label{fig:resnetattention}
    \end{figure}
\FloatBarrier

\chapter{Experimental Results}

We implemented a range of preprocessing measures to overcome constraints related to computational resources and harmonize our data. The MRI images in the dataset have the size of 240x240x155, which were cropped to a size of 128x128x128. The reduced dimensionality makes the computational process more feasible and minimizes the risk of overfitting by reducing the complexity of our model. Further preprocessing involved the normalization of image intensities via min-max normalization. This process standardized the intensity range between -1 and 1 across all images, allowing the model to focus on the significant features and reduce potential bias due to different intensity levels.
Moreover, we concatenated all four modalities for each patient to utilize the maximum available information from the scans, creating a 4-channel input. This approach ensures that the model exploits the comprehensive information available in these sequential datasets, allowing us to leverage each MRI modality's unique features and variations. As a result, the model was trained on data with a dimension of 128x128x128x4, with the fourth dimension representing the four modalities. The model has approximately 3 million parameters and 86.58 GFlops. Also, we used the flipping augmentation technique to increase the dataset.
The choice of the BraTS 2020 dataset for this study was motivated by the quality and diversity of the data it provides. Its public availability and widespread use in the research community further ensure the reproducibility and comparability of our findings. In addition to BraTS2020, we tested our model on BraTS 2019 to evaluate its performance and adaptability to different data. Although this dataset provides less data in compared to BraTS2020, the results highlight our model's ability to effectively segment whole tumors in compared with the current methods.
Our methodology, built using Python programming language, leverages the capabilities of the PyTorch library ~\cite{paszke2019pytorch}. The optimization process utilized the ADAM optimizer, beginning with a learning rate of 5e-4. To further enhance the training, a learning rate scheduler based on the ReduceLROnPlateau function with patience of 4 was employed. This effectively adjusted the learning rate downwards whenever the model's performance reached a plateau. In terms of architectural composition, we utilized the ReLU activation function in conjunction with group normalization. This combination has been proven to enhance our model's stability and performance by normalizing the network. In spite of the constraints related to computational resources, we succeeded in training the model over a cumulative of 200 epochs, using a batch size of 4. Additionally, the input image was constructed using four modalities stacked together, each with a size of 128*128*128. The data was divided so that 80\% was allocated for training, 10\% for testing, and 10\% for validation.
In the realm of brain tumor analysis, it's vital to acknowledge the intrinsic challenge posed by imbalanced classes within the datasets. These imbalances can occur when certain types of tumor subregions, such as Necrosis and non-enhancing tumor (NET), peritumoral edema (ED), or enhancing tumor (ET), are underrepresented or overrepresented relative to others. Such disparities can significantly skew the model's learning process, leading to biased predictions and potentially undermining the model's clinical utility. To counteract this, dividing the datasets to address imbalanced classes is an essential step that ensures a more effective and equitable distribution of data. By stratifying the dataset based on the prevalence of each tumor subregion, researchers can create more balanced datasets that accurately reflect the variety of tumor characteristics encountered in clinical practice. Therefore, the correlations between these subregions and the overall tumor size are evaluated for each patient. The correlation coefficient between NET and tumor size is approximately 0.58, suggesting a moderate positive correlation. This means that as the tumor size increases, the volume of the necrosis and non-enhancing tumor tends to increase as well, but not as strongly as the other components. The correlation between ED and tumor size is approximately 0.81, which indicates a strong positive correlation. This is visible in the scatter plot showing that patients with larger tumors tend to have more extensive peritumoral edema. This could suggest that as the tumor grows, it causes more inflammation and disruption to the surrounding brain tissue, resulting in more edema. The correlation coefficient between ET and tumor size is approximately 0.49, indicating a moderate positive correlation, although it is weaker than the correlation between ED and tumor size. This plot shows that while there is a trend for enhancing tumors to be larger when the overall tumor size is bigger, the relationship is not as consistent or strong as with ED. This could imply that other factors besides overall tumor size might influence the volume of the enhancing tumor.

\FloatBarrier
    \begin{figure}[htbp]
    \centering
     \includegraphics[width=\textwidth, keepaspectratio]{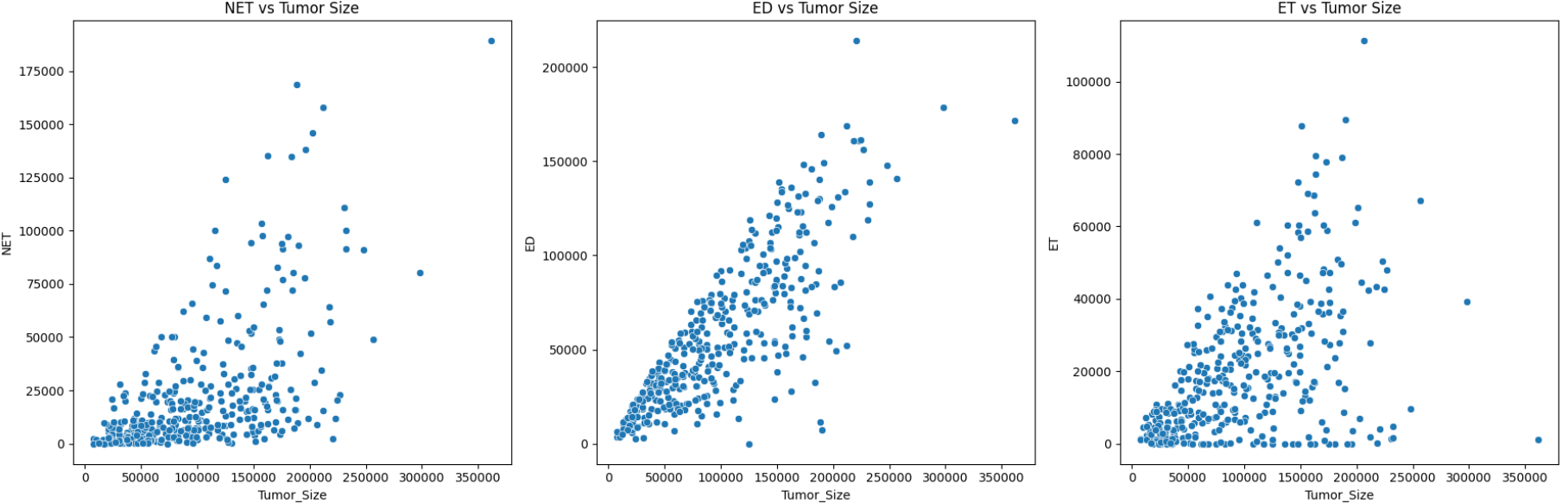}
    \caption[The correlation between these sub-regions and the overall tumor size]{Scatter plots illustrating the correlation between the size of various tumor sub-regions (NET, ED, and ET) and the overall tumor volume. The graphs indicate varying degrees of association}
    \label{fig:tumorsize}
    \end{figure}
\FloatBarrier

Given the varying correlation strengths between tumor size and the subregions of NET, ED, and ET, it is clear that these metrics reflect different aspects of the tumor's biology. Stratification based on these features ensures that each set contains a representative distribution of these characteristics, which is crucial for the generalization of the model and its consistency in evaluation. Without this careful distribution, the model might perform well on the test set simply because the test set has an unrepresentative distribution of easy or hard cases based on tumor composition.

To verify the efficiency and enhancement in the segmentation performance of our developed model, we employed five-fold cross-validation. Cross-validation is a robust statistical technique used to evaluate the performance and stability of predictive models. Fig 5.2 illustrates a comprehensive stratified five-fold cross-validation scheme. The delineation into training, validation, and test sets for each fold ensures that each subset not only receives diverse instances from the dataset but also maintains a proportional distribution of key tumor features (NET, ED, and ET).
\FloatBarrier
    \begin{figure}[htbp]
    \centering
     \includegraphics[width=\textwidth, keepaspectratio]{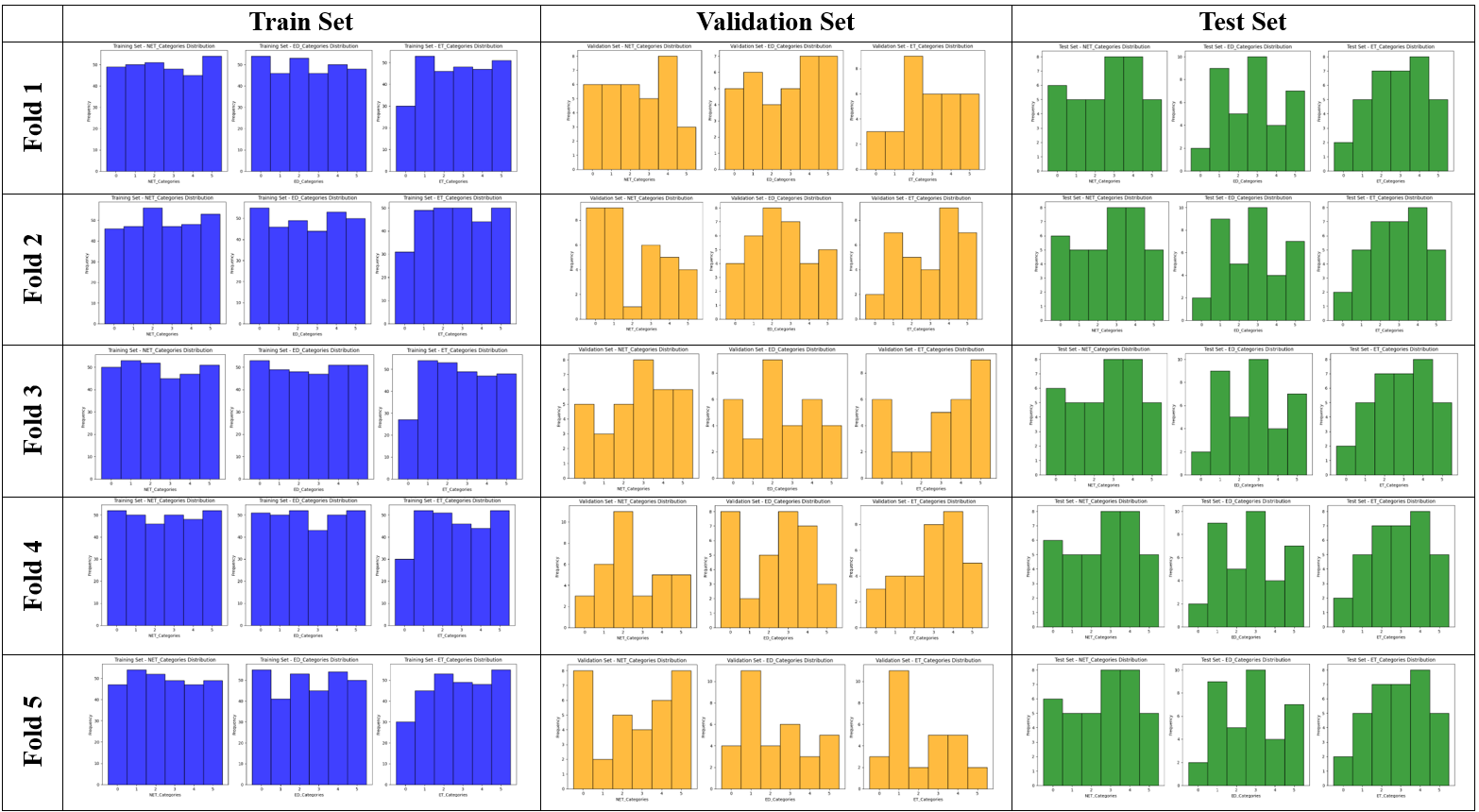}
    \caption[The distribution of key tumor features (NET, ED, and ET) on Train, Validation and Test sets]{Bar charts representing the distribution of key tumor features (NET, ED, and ET) across training, validation, and test datasets for each of the five folds in a cross-validation framework.}
    \label{fig:fivefold}
    \end{figure}
\FloatBarrier

\section*{Evaluation measures}

\subsection*{A. Dice Coefficient Score}
The Dice Coefficient Score (DSC) is a metric employed to assess the likeness between two sets. Within the scope of image segmentation, for example, in brain tumor segmentation, these "sets" indicate the anticipated segmentation and the actual or 'ground truth' segmentation.

\noindent DSC is particularly beneficial in medical imaging because it provides a measure of overlap that is easy to interpret. It evaluates how close the predicted segmentation is to the actual segmentation by quantifying the spatial overlap accuracy. Therefore, a higher DSC suggests a higher match between the predicted and actual values, leading to better segmentation results.

\noindent The DSC Equation is calculated as follows:
\begin{equation}
DSC = \frac{2 \times |X \cap Y|}{|X| + |Y|}
\end{equation}
In this context, \(X\) represents the anticipated segmentation, while \(Y\) stands for the actual truth. The symbol \(\cap\) implies intersection, and \(|\cdot|\) signifies the magnitude of the set. Each voxel of the tumor is tagged as 1, and those not part of the tumor are tagged as 0.

\subsection*{B. Hausdorff Distance}
The Hausdorff Distance (HD) is a well-liked metric for assessing the similarity between two sets of points ~\cite{rucklidge1996efficient}. It essentially measures the maximum distance between any point in one set and its closest corresponding point in the other set. Lower Hausdorff distance indicates a closer match between the automated segmentation and the ground truth, signifying better segmentation accuracy and higher Hausdorff distance implies a larger discrepancy between the boundaries, suggesting potential errors in the segmentation. The Hausdorff distance \(d_H(X, Y)\) is a metric used to measure the extent to which two sets \(X\) and \(Y\) differ. It is defined as:

\begin{equation}
d_H(X, Y) = \max \left\{ \sup_{x \in X} \inf_{y \in Y} d(x, y), \sup_{y \in Y} \inf_{x \in X} d(y, x) \right\}
\end{equation}

Where \(d(x, y)\) is the distance between points \(x\) and \(y\), \(\sup\) represents the supremum, capturing the largest of the minimal distances from points in one set to points in the other, and \(\inf\) denotes the infimum, the smallest distance from a point to any point in the other set. This measure is especially useful in applications such as image segmentation, where it helps in evaluating the similarity of shapes.

\subsection*{C. Sensitivity}
Sensitivity measures the proportion of actual positives that are correctly identified by the model. In the context of brain tumor segmentation, it quantifies how effectively the model identifies tumor voxels. High sensitivity minimizes the risk of overlooking potential tumors. The equation for sensitivity is:

\begin{equation}
\text{Sensitivity} = \frac{TP}{TP + FN}
\end{equation}
Where TP (True Positives) is the number of tumor voxels correctly identified as tumor and FN (False Negatives) is the number of tumor voxels incorrectly identified as non-tumor.

\subsection*{C. Specificity}
Specificity measures the proportion of actual negatives that are correctly identified by the model. In brain tumor segmentation, it evaluates how well the model identifies non-tumor voxels. Specificity is important to ensure that the model is not overly sensitive and marking too many areas as tumors. The equation for specificity is:
\begin{equation}
\text{Specificity} = \frac{TN}{TN + FP}
\end{equation}
Where TN (True Negatives) is the number of non-tumor voxels correctly identified as non-tumor and FP (False Positives) is the number of non-tumor voxels incorrectly identified as tumor.

\subsection*{D. Statistical Significance (Paired t-test)}

The paired t-test is utilized to ascertain whether the mean difference between two sets of paired observations significantly deviates from zero. The formula for computing the t-statistic in a paired t-test is expressed as:

\[
t = \frac{\bar{d}}{s_{\bar{d}}}
\]

where: $\bar{d}$ denotes the mean of the differences between the paired observations. This represents the average change between the two sets of paired data (the differences in Dice scores between the ADRUwAMS model and the base 3D U-Net for each fold cross validation). $s_{\bar{d}}$ is the standard error of the mean difference, calculated as $\frac{s_{d}}{\sqrt{n}}$, where: $s_{d}$ is the standard deviation of the differences, quantifying the dispersion or spread of these differences. $n$ signifies the number of paired observations (folds). The p-value obtained from the t-statistic allows us to assess the statistical significance of the observed differences:

\begin{itemize}
  \item A \textbf{low p-value} (typically $< 0.05$) suggests that the observed difference in means is unlikely to have occurred by chance, indicating statistically significant differences between the two models.
  \item A \textbf{high p-value} ($\geq 0.05$) implies that the difference in means could likely be due to random sampling variability, and hence, is not statistically significant.
\end{itemize}

The determination of statistical significance is fundamental in concluding whether the improvements or changes observed with the ADRUwAMS model are reliable and not results of random fluctuations in the data.

\subsection*{E. Effect Size (Cohen's d)}

While the paired t-test tells if there is a statistically significant difference, Cohen's d measures how large that difference is. Cohen's d is a standardized effect size that helps understand the practical impact of the difference, regardless of the sample size, providing a sense of the magnitude of the improvement. Cohen's d is computed using the following expression:

\[
d = \frac{\bar{d}}{s_{d}}
\]

where $\bar{d}$ is identical to the mean difference used in the t-test. $s_{d}$ represents the standard deviation of the differences. This is the effect size measure that indicates the size of the difference relative to the variability observed in the data. Cohen's d values are typically interpreted as follows:

\begin{itemize}
  \item \textbf{0.2}: Denotes a small effect size.
  \item \textbf{0.5}: Indicates a medium effect size.
  \item \textbf{0.8}: Signifies a large effect size.
\end{itemize}

The larger the absolute value of Cohen's d, the larger the effect size, indicating a more substantial difference between the two groups being compared. it allows us to understand the magnitude of the difference in performance between the ADRUwAMS model and the base 3D U-Net.

\section{Experimental Results}
In this section, we present and discuss the findings from our experiment on optimizing a deep learning model for medical image segmentation, specifically focusing on brain tumor segmentation. Our evaluation centers around four critical metrics: the Dice Coefficient Score (DSC), the Hausdorff Distance, Sensitivity, and Specificity, which collectively provide a comprehensive assessment of the model's accuracy and reliability. We assess the model over 200 epochs, analyzing the trends in loss values. In Figure 5.3, we observe the model's performance across the 200 epochs, focusing on the loss values for the training and validation datasets. A key observation is the gradual reduction in loss values over these epochs, suggesting that the model is learning effectively and improving in its segmentation accuracy. Fig 5.4 shows the model's performance on the segmentation of brain tumors.

\FloatBarrier
    \begin{figure}[htbp]
    \centering
     \includegraphics[width=0.8\textwidth, keepaspectratio]{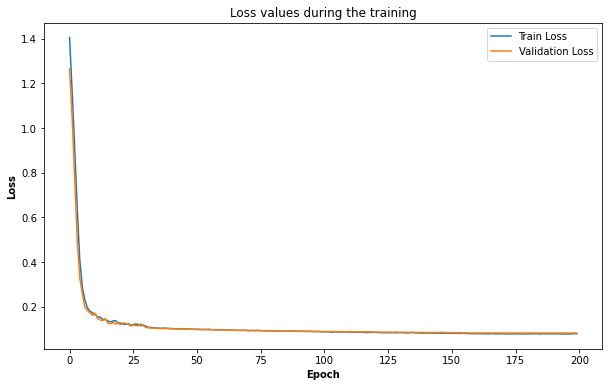}
    \caption[Loss values during the training]{Training and Validation Loss Trends over 200 Epochs. This graph tracks the decline in loss values for both the training and validation sets, indicating the model's progressive learning and increasing accuracy in brain tumor segmentation over time.}
    \label{fig:loss}
    \end{figure}
\FloatBarrier

The Dice Score is utilized to gauge the overlap between two instances. A higher Dice Score indicates better overlap, suggesting the better performance of the model in accurately segmenting the tumor area. Regarding the Dice Score, our proposed model provides a good performance in segmenting Whole Tumor (WT), Tumor Core (TC), and Enhanced Tumor (ET), attaining scores of 0.92, 0.84, and 0.80, respectively. Also, we used HD to provide an estimate of the worst-case segmentation error. Lower values imply better performance in terms of segmentation accuracy. Our proposed model ADRUwAMS demonstrates a better average performance compared to all the other models, with HD scores for Whole Tumor, Tumor Code, and Enhanced Tumor being 1.32, 3.04, and 10.52, respectively. Table 5.1 presents a comparative analysis between the ADRUwAMS model and the state-of-the-art models using the BraTS 2020 dataset. The bolded values show the model's performance in comparison to the other methods. Table 5.2 shows the module's performance using five-fold cross-validation. Also, the proposed method has been tested on BraTS 2019, which is depicted in Table 5.4.

A further evaluation of the ADRUwAMS model is compared with a baseline established by a simple 3D U-Net model, focusing on key metrics: Dice Score and Hausdorff Distance (HD) across three tumor regions: Whole Tumor (WT), Tumor Core (TC), and Enhancing Tumor (ET). Both models underwent 5-fold cross-validation, and performance metrics were recorded for each fold. The analysis includes statistical tests to assess significance and effect size measurements to gauge the magnitude of differences. Table 5.3 shows five-fold cross validation of 3D-unet as base line.
The paired t-test results demonstrate the statistical significance of the improvements that ADRUwAMS 3D U-Net model has achieved over the simple 3D U-Net across different metrics and tumor classes. For Dice scores, highly significant p-values were observed for Whole Tumor (WT, p=0.000132), Tumor Core (TC, p=6.61e-05), and Enhancing Tumor (ET, p=0.000902) classes, confirming the enhancements are statistically substantial. In terms of Hausdorff Distance (HD), the improvements were also statistically significant for WT (p=0.000871) and TC (p=0.001174) classes, although the ET class improvements did not reach conventional significance (p=0.060994), suggesting a potential trend that needs further investigation.
The Cohen's d metrics indicate large effect sizes for both Dice scores and HD metrics in most comparisons, signifying the practical importance and magnitude of the improvements. Specifically, the Dice score enhancements exhibit large effect sizes for WT (6.92), TC (9.96), and ET (6.32), indicating marked improvements. For HD metrics, the large negative effect sizes for WT (-3.991) and TC (-3.692) highlight substantial model advancements in boundary delineation accuracy, with a moderate effect size for ET (-1.156), suggesting noticeable but less pronounced improvements.
The substantial statistical enhancements observed with the ADRUwAMS 3D U-Net model have significant real-world implications, especially in the medical field where precise tumor segmentation is crucial for effective treatment planning and outcome assessment.

    \begin{figure}[H]
    \centering
     \includegraphics[width=0.78\textwidth, keepaspectratio]{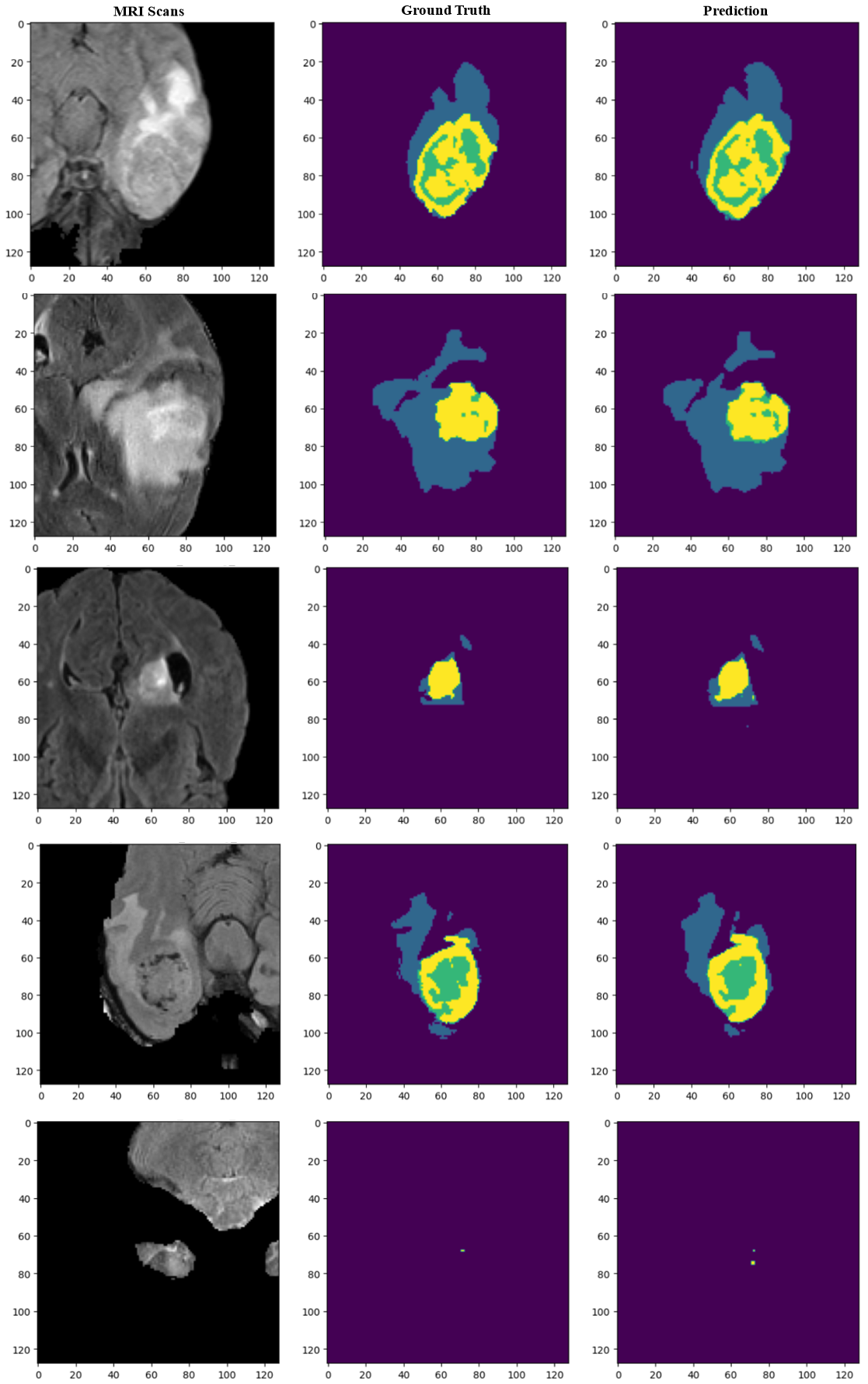}
    \caption[The Model's performance on segmentation of Brain tumor.]{Evaluation of Segmentation Accuracy on Brain Tumors. The first column displays MRI scans, the middle column shows the Ground Truth for tumor areas, and the third column presents the model's Predictions. The comparison across columns illustrates the model's proficiency in segmenting different tumor regions against the established Ground Truth.}
    \label{fig:result}
    \end{figure}

\begin{table}[t]
\centering
\caption{Performance comparison of different architectures.}
\label{tab:performance_comparison}
\begin{adjustwidth}{-2.5cm}{-2cm}
\small
\begin{tabular}{@{}p{5cm}ccccccccc@{}}
\toprule
\textbf{Architecture} & \multicolumn{3}{c}{\textbf{Dice Score}} & \multicolumn{3}{c}{\textbf{HD}} \\
\cmidrule(lr){2-4} \cmidrule(l){5-7}
                       & \textbf{WT}    & \textbf{TC}   & \textbf{ET}   & \textbf{WT}    & \textbf{TC}   & \textbf{ET}   \\
\midrule
Dual-Path attention U-Net  ~\cite{jun2021brain} & 0.8780$_{\pm0.112}$ & 0.7790$_{\pm0.199}$ & 0.7520$_{\pm0.282}$ & 6.3000$_{\pm10.03}$ & 11.0200$_{\pm34.49}$ & 30.6500$_{\pm96.09}$ \\
3D self-ensemble ResUNet  ~\cite{pei2021multimodal}   & 0.89 & 0.81 & 0.76 & 5.28 & 7.74 & 33.26 \\
Modified 3D U-net  ~\cite{parmar2021brain}  & 0.90$_{\pm0.111}$ & 0.84$_{\pm0.0.257}$ & 0.74$_{\pm0.0.2158}$ & 5.08$_{\pm5.988}$ & 8.69$_{\pm79.565}$ & 36.04$_{\pm074.8080}$ \\
TransBTS ~\cite{wang2021transbts}  & 0.8900 & 0.8136 & 0.7850 & 6.4690 & 10.4680 & 16.7160 \\
MENet  ~\cite{zhang2021me}  & 0.8800 & 0.7400 & 0.7000 & 6.9500 & 30.1800 & 38.6000 \\
Ensembles of CNN  ~\cite{gonzalez20213d}  & 0.902$_{\pm0.066}$ & 0.815$_{\pm0.161}$ & 0.773$_{\pm0.257}$ & 6.16$_{\pm11.2}$ & 7.55$_{\pm11.4}$ & 21.80$_{\pm79}$ \\
Swinbts  ~\cite{jiang2022swinbts}  & 0.8906$_{\pm0.130}$ & 0.8030$_{\pm0.079}$ & 0.7736$_{\pm0.224}$ & 8.56 & 15.78 & 26.84 \\
3D PSwinbts  ~\cite{liang20223d}  & 0.9076$_{\pm1}$ & 0.8420$_{\pm3.6}$ & 0.7948$_{\pm4.8}$ & 5.573$_{\pm0.81}$ & 7.252$_{\pm6.01}$ & 19.437$_{\pm14.891}$ \\
Pre-operative three-dimensional MRI  ~\cite{kaur2023deep} & 0.88 & 0.79 & 0.77 & 7.79 & 13.86 & 32.69 \\
HMNet  ~\cite{zhang2023hmnet}  & 0.901 & 0.823 & 0.781 & 5.954 & 7.055 & 21.340 \\
Segmentation via pixel-level and feature-level image fusion  ~\cite{liu2022brain} & 0.8950 & 0.8178 & 0.7745 & 5.3117 & 9.4285 & 4.4715 \\
dResU-Net  ~\cite{raza2023dresu} &0.8660&0.8357&0.8004&-&-&- \\
Selective Deeply Supervised Multi-Scale Attention Network  ~\cite{rehman2023selective} & 0.9024$_{\pm7.21}$ & 0.8693$_{\pm10.58}$ & 0.8064$_{\pm18.46}$ & 4.27$_{\pm9.56}$ & 6.32$_{\pm9.88}$ & 5.87$_{\pm22.13}$ \\
MM-UNet  ~\cite{zhao2022mm} &0.8500&0.7650&0.7620&8.243&10.766&6.389 \\
ADRUwAMS &0.9229$_{\pm0.004}$&0.8432$_{\pm0.008}$&0.8004$_{\pm0.018}$ &1.3228$_{\pm0.238}$&3.0375$_{\pm0.168}$ &10.5287$_{\pm3.422}$ \\
\bottomrule
\end{tabular}
\end{adjustwidth}
\end{table}

\begin{table}[t]
\centering
\caption{Quantitative results of the proposed method on brain tumor segmentation.}
\label{tab:quantitative_results_proposed}
\small
\setlength\tabcolsep{2.5pt}
\begin{adjustwidth}{-1.5cm}{-1.5cm}
\begin{tabular}{@{}lcccccccccccc@{}}
\toprule
Method & \multicolumn{3}{c}{\textbf{Dice Score}} & \multicolumn{3}{c}{\textbf{HD}} & \multicolumn{3}{c}{\textbf{Sensitivity}} & \multicolumn{3}{c}{\textbf{Specificity}} \\
\cmidrule(lr){2-4} \cmidrule(lr){5-7} \cmidrule(lr){8-10} \cmidrule(l){11-13}
& \textbf{WT} & \textbf{TC} & \textbf{ET} & \textbf{WT} & \textbf{TC} & \textbf{ET} & \textbf{WT} & \textbf{TC} & \textbf{ET} & \textbf{WT} & \textbf{TC} & \textbf{ET} \\
\midrule
Fold 1 & 0.9216 & 0.8506 & 0.8219 & 1.4021 & 2.8624 & 6.9510 & 0.9312 & 0.8848 & 0.8449 & 0.9191 & 0.8640 & 0.8458 \\
Fold 2 & 0.9212 & 0.8435 & 0.8204 & 1.5128 & 3.0682 & 6.9243 & 0.9214 & 0.9013 & 0.8421 & 0.9269 & 0.8393 & 0.8500 \\
Fold 3 & 0.9314 & 0.8386 & 0.7730 & 0.8930 & 3.2084 & 16.1447 & 0.9338 & 0.8942 & 0.8476 & 0.9320 & 0.8416 & 0.8003 \\
Fold 4 & 0.9231 & 0.8534 & 0.7925 & 1.2530 & 2.8359 & 11.3218 & 0.9322 & 0.8903 & 0.8276 & 0.9216 & 0.8604 & 0.8391 \\
Fold 5 & 0.9172 & 0.8284 & 0.7963 & 1.5535 & 3.2397 & 11.3021 & 0.9172 & 0.8689 & 0.8454 & 0.9270 & 0.8471 & 0.8247 \\
Mean  & 0.9229 & 0.8432 & 0.8008 & 1.3228 & 3.0429 & 10.5287 & 0.9271 & 0.8879 & 0.8415 & 0.9253 & 0.8504 & 0.8319 \\
SD      & 0.0046 & 0.0089 & 0.0183 & 0.2388 & 0.1686 & 3.4222 & 0.0066 & 0.0109 & 0.0071 & 0.0045 & 0.0099 &0.0180 \\
\bottomrule
\end{tabular}
\end{adjustwidth}
\end{table}

\begin{table}[t]
\centering
\caption{Quantitative results of the 3D-Unet Model (base model) on brain tumor segmentation.}
\label{tab:quantitative_results_baseline}
\small
\setlength\tabcolsep{2.5pt}
\begin{adjustwidth}{-1.5cm}{-1.5cm}
\begin{tabular}{@{}lcccccccccccc@{}}
\toprule
Method & \multicolumn{3}{c}{\textbf{Dice Score}} & \multicolumn{3}{c}{\textbf{HD}} & \multicolumn{3}{c}{\textbf{Sensitivity}} & \multicolumn{3}{c}{\textbf{Specificity}} \\
\cmidrule(lr){2-4} \cmidrule(lr){5-7} \cmidrule(lr){8-10} \cmidrule(l){11-13}
& \textbf{WT} & \textbf{TC} & \textbf{ET} & \textbf{WT} & \textbf{TC} & \textbf{ET} & \textbf{WT} & \textbf{TC} & \textbf{ET} & \textbf{WT} & \textbf{TC} & \textbf{ET} \\
\midrule
Fold 1 & 0.8897 & 0.7593 & 0.6994 & 2.3118 & 4.7868 & 17.4723 & 0.9109 & 0.8129 & 0.7755 & 0.8830 & 0.7931 & 0.7593 \\
Fold 2 & 0.8947 & 0.7608 & 0.6995 & 1.9869 & 4.8476 & 17.6157 & 0.8964 & 0.7864 & 0.7690 & 0.9014 & 0.8063 & 0.7623 \\
Fold 3 & 0.8953 & 0.7632 & 0.7105 & 1.8619 & 4.2071 & 13.3497 & 0.8994 & 0.8056 & 0.7714 & 0.9006 & 0.7841 & 0.7626 \\
Fold 4 & 0.8968 & 0.7490 & 0.6899 & 2.0270 & 5.0622 & 18.0932 & 0.9135 & 0.8335 & 0.7876 & 0.8932 & 0.7554 & 0.7370 \\
Fold 5 & 0.8914 & 0.7472 & 0.7090 & 2.2766 & 5.4035 & 17.7841 & 0.8997 & 0.7823 & 0.7732 & 0.9007 & 0.8036 & 0.7740 \\
Mean  & 0.8935 & 0.7559 & 0.7016 & 2.0928 & 4.8614 & 16.8630 & 0.9039 & 0.8041 & 0.7753 & 0.8957 & 0.7885 & 0.7590 \\
SD    & 0.0026 & 0.0065 & 0.0074 & 0.1735 & 0.3919 &1.7687 & 0.0068 & 0.0186 & 0.0064 & 0.0070 & 0.0183& 0.0121 \\
\bottomrule
\end{tabular}
\end{adjustwidth}
\end{table}

\begin{table}[t]
\centering
\caption{Comparison of the proposed model with other approaches on the BraTS 2019 dataset}
\label{tab:performance_comparison2019}
\begin{adjustwidth}{-2.5cm}{-2cm}
\small
\begin{tabular}{@{}p{5cm}ccccccccc@{}}
\toprule
\textbf{Architecture} & \multicolumn{3}{c}{\textbf{Dice Score}} & \multicolumn{3}{c}{\textbf{HD}} \\
\cmidrule(lr){2-4} \cmidrule(l){5-7}
                      & \textbf{WT}    & \textbf{TC}   & \textbf{ET}   & \textbf{WT}    & \textbf{TC}   & \textbf{ET}   \\
\midrule
dual supervision guided attentional network ~\cite{zhou2022dual} & 0.882 & 0.771 & 0.727 & 8.09 & 10.3 & 6.6 \\
heuristic approach for segmentation  ~\cite{di2021application} & 0.8598 & 0.7728 & 0.7153 & - & - & - \\
RAAGR2-Net ~\cite{rehman2023raagr2} & 0.884 & 0.814 & 0.763 & - & - & - \\
Swinbts ~\cite{jiang2022swinbts} & 0.8975$_{\pm0.070}$ & 0.7928$_{\pm0.232}$ & 0.7443$_{\pm0.294}$ & - & - & - \\
Cascaded 3D U-Net and 3D U-Net++ ~\cite{li2022automatic} & 0.867 & 0.834 & 0.802 & - & - & - \\
Multiscale lightweight 3D segmentation with attention mechanism ~\cite{liu2023multiscale} & 0.8994 & 0.8349 & 0.7791 & 5.45 & 6.56 & 4.03 \\
Proposed Method & 0.9060$_{\pm0.002}$ & 0.8279$_{\pm0.007}$ & 0.72939$_{\pm0.009}$ & 2.0552$_{\pm0.210}$ & 3.2217$_{\pm0.667}$ & 23.7989$_{\pm0.673}$ \\
\bottomrule
\end{tabular}
\end{adjustwidth}
\end{table}

\chapter{Conclusion and Future Works}

\section{Discussion}
Brain tumor segmentation is an essential task in medical diagnostics. Recently, U-Net models have shown significant promise in these complex automatic segmentation tasks. However, there are computational limitations when dealing with high-resolution 3D images like brain MRIs. Our study proposed an innovative Adaptive Dual Residual U-Net with Attention Gate and Multiscale Spatial Attention Mechanisms (ADRUwAMS). The novelty of our approach lies in the architecture of residual networks along with group normalization and incorporation of these adaptive residual networks with attention gates and multiscale spatial attention mechanisms. The dual residual network helps the model learn features at different levels of abstraction, capturing the intricate features of gliomas. The attention gate mechanism reduces the impact of irrelevant regions, allowing the model to focus on the more informative tumor regions. The multiscale spatial attention mechanism maintains the spatial context of images, ensuring the accurate segmentation of smaller or hard-to-detect tumor regions. This approach has made an advancement in brain tumor segmentation by enhancing segmentation performance and surpassing previous methods in terms of precision and proficiency.
The ADRUwAMS model's notable advancements in Dice scores and Hausdorff Distances for brain tumor segmentation directly translate to significant clinical benefits. Enhanced accuracy in defining the Whole Tumor (WT) and Tumor Core (TC) regions promises improved precision in medical analyses. Such precision, underscored by large effect sizes, indicates meaningful gains in clinical applications, suggesting that these statistical enhancements could lead to more reliable assessments and interventions. Although the Enhancing Tumor (ET) area's improvements were less pronounced in statistical terms, the observed trends and moderate effect sizes still point to potential gains in clinical specificity and treatment monitoring. These incremental yet impactful enhancements highlight the ADRUwAMS model's potential to refine diagnostic processes, marking a step forward in personalized patient care and treatment efficacy.
While our novel ADRUwAMS model has shown promising results in detecting brain tumors, there are certain limitations that need to be addressed. First, the model employs the same convolutional operations in both the dual residual network and the attention mechanisms. This approach might not sufficiently consider the distinct computational requirements necessary to effectively handle the intricate low-level image details and the more complex high-level features. Furthermore, Limited datasets might reduce the model's robustness and generalizability in detecting a wide range of brain tumors. For future work, we aim to enhance the dual residual network structure to provide more efficiency to the varying nature of the extracted features. Additionally, we intend to explore advanced training techniques to optimize the model's performance even with limited datasets, consequently improving its tumor detection and segmentation efficacy.
Looking toward the future, the field of tumor segmentation presents several promising directions. The incorporation of temporal information from longitudinal studies, for example, could provide further insights into tumor growth patterns. Extending the model to use multimodal imaging data can provide complementary information about tumors, thereby enhancing model robustness. Also, perfusion MRI, which offers insights into the blood flow within the tumor, could significantly augment the model's diagnostic capabilities, offering new avenues for understanding tumor heterogeneity and vascular characteristics. The adoption of generative models, like Generative Adversarial Networks (GANs), can address the common challenge of limited labeled data availability by generating synthetic tumor images for training. Similarly, transfer learning can be a potent strategy to bolster model performance, especially in cases where training data for brain tumors is scarce. With future refinements and the exploration of promising opportunities, we aim to persistently build on this progress and contribute to enhancing segmentation in the field of medical imaging.

\section{Conclusion}
Our proposed model, with its dual residual network structure, successfully captures both high-level semantic features and low-level details from brain images. The comprehensive feature learning allows the model to effectively distinguish between hard regions, tumor regions, and healthy tissues, demonstrating improved precision and accuracy in isolating different tumor types and parts. The integration of the attention mechanism assigns varying weights to different image regions, enabling the model to concentrate on hard regions while lessening the impact of non-tumor areas. This mechanism improves overall segmentation performance and enhances boundary delineation. Furthermore, the spatial attention mechanism is crucial in preserving the spatial context of images. It maintains intricate spatial structures, facilitating the accurate segmentation of smaller or hard regions of the tumor, which might otherwise be overlooked or lost during the segmentation process. Our experimental results demonstrate the proposed ADRUwAMS method outperforms contemporary state-of-the-art methods.

\cleardoublepage
\phantomsection
\addcontentsline{toc}{chapter}{References}
\setlength{\itemsep}{0.3em}

\end{document}